\documentclass{article}
\usepackage{log_2026}						

\usepackage{booktabs}						
\usepackage{multirow}						
\usepackage{amsfonts}						
\usepackage{graphicx}						
\usepackage{duckuments}						
\usepackage{subcaption}
\usepackage[numbers,compress,sort]{natbib}	

\title[KGCache:  Amortized Subgraph Retrieval for KG Reasoning with LLMs \logyear]{KGCache:  Amortized Subgraph Retrieval for KG Reasoning with LLMs}

\author[Stanic et al.]{%
Uros Stanic\\
Texas A\&M University \\
\email{stanic@tamu.edu}\And
Changcheng Yuan\\
Texas A\&M University\\
\email{ericycc@tamu.edu}\And
Sabuj Laskar\\
Texas A\&M University\\
\email{sabuj.laskar@tamu.edu}\And
Ariful Azad\\
Texas A\&M University\\
\email{ariful@tamu.edu}
}

\begin{document}

\maketitle

\begin{abstract}
Large language models can answer knowledge-intensive questions more reliably when they are grounded with knowledge graphs, but systems such as Think-on-Graph and Reasoning-on-Graph repeatedly query the same graph neighborhoods across different questions. In this work, we study this repeated retrieval in Knowledge Graph Question Answering~(KGQA) workloads and propose KGCache, an in-memory cache for one-hop knowledge graph neighborhoods. KGCache is designed to be compatible with both iterative traversal (ToG) and one shot planning (RoG) KGQA paradigms. KGCache is placed between the KGQA engine and the backend serving the KG, so repeated entity requests can be served from cache instead of issuing new KG queries. We evaluate KGCache on WebQSP and CWQ using LRU, LFU, and a trace-aware Oracle policy. Our analysis shows that both datasets contain substantial entity reuse among starting entities and entities reached during traversal. We also explore semantic caching for similar queries, which shows additional hit-rate gains on WebQSP and needs further accuracy testing on CWQ. Entity caching accelerates KG retrieval by up to $1.91\times$, while semantic-context caching achieves up to $1.08\times$ full-system speedup in the evaluated configurations, with each hit being up to $17.98\times$ faster. End-to-end gains are bounded by the share of runtime spent on KG retrieval, so caching is most beneficial when KG access is slow, e.g., with remote or large-scale KG backends.
\end{abstract}

\section{Introduction}
\label{sec:intro}
Large language models (LLMs) have remarkable capabilities across a broad range of natural language understanding and generation tasks, yet they remain fundamentally constrained by their reliance on static parametric knowledge acquired during pretraining \cite{KGclaim2}. This limitation manifests acutely in knowledge-intensive question answering, where models are prone to hallucination, generating fluent but factually incorrect responses, particularly when questions require up-to-date or multi-hop reasoning over structured knowledge \cite{KGclaim}. 
Knowledge graphs (KGs) offer an elegant solution.
By encoding facts as structured (subject, relation, object) triples over large entity sets, KGs provide a verifiable, traversable, and updatable external memory that can ground LLM reasoning in factual evidence \cite{KGclaim}.

Recent work combines LLMs with KGs through Knowledge Graph Question Answering (KGQA)~\cite{TOG_01}. In this setting, an LLM-guided system retrieves KG evidence and uses it to answer a question. 
Think-on-Graph (ToG)~\cite{TOG_01} performs this process iteratively, leveraging the LLM at each hop to choose relations and entities that expand the search frontier.
ToG and its successors show that such KG retrieval can improve answer grounding on standard benchmarks \cite{TOG_02, TOG_03}. This design, however, invokes the LLM and queries the KG several times per question. Reasoning-on-Graph (RoG) instead generates a relation-path plan before KG execution, reducing planning-stage LLM invocations while maintaining competitive answer quality \cite{rog}. RoG still executes structured KG queries, and different questions can revisit the same graph regions.

A source of avoidable work in this pipeline is \textit{redundant subgraph retrieval}. KGQA benchmark workloads can repeatedly access the same entities, such as people, countries, and institutions. 
For example, 75.9\% and 44.7\% of entities recur across {CWQ}~\cite{CQW} and {WebQSP}~\cite{WebQSP} benchmarks, respectively, indicating that accesses are concentrated among a subset of entities.
Redundant lookups in the backend KG store incur significant overhead, as each neighborhood query involves structured query execution (\textit{e.g.,} SPARQL), data transfer, and result serialization~\cite{KGclaim}.
Prior GraphRAG research has emphasized retrieval \textit{quality}, including subgraph selection, relation selection, and evidence construction~\cite{query_aware_gnn, cai2025simgrag}, but cross-query amortization of KG access has received comparatively less attention. 
The opportunity is most direct in ToG-style workloads because they issue multiple neighborhood requests per question, although RoG path execution can also revisit entities across questions.

We present \textbf{KGCache}, an \textbf{entity-level caching} layer for KG-augmented LLM reasoning that amortizes repeated subgraph retrieval. 
KGCache is placed between the reasoning engine and the KG backend and maps entity identifiers to one-hop neighborhoods. For a fixed KG snapshot, a repeated request can be served from memory without querying the backend. 
This entity-level caching accelerates KG lookups but leaves the LLM call sequence unchanged, since each traversal step still invokes the reasoning engine.
To also reduce LLM calls, we develop a complementary \textbf{semantic-context cache} that operates one level above entity caching. 
When an incoming query is sufficiently similar to a previously seen query, measured by embedding similarity, the semantic cache returns the associated KG context directly, skipping both backend invocation and the KG traversal steps. 

We evaluate several cache management policies, including Least Recently Used (LRU), Least Frequently Used (LFU), and a trace-aware Oracle. We integrate caching with multiple KG backends and LLM frontends and show that it reduces both KG traversals and the number of LLM calls in the TOG workload across two KGQA benchmarks. In particular, entity-level caching reduces backend queries by up to $38\%$, yielding a $1.9\times$ speedup in KG retrieval steps. 
Our contribution is not a new cache replacement algorithm. Instead, we show that classical caching applied at the KG-backend boundary is an effective, low-overhead optimization for KG-augmented LLM inference, and we characterize when it pays off. The main contributions of this paper are as follows:
\begin{enumerate}
    \item {\bf Workload Characterization:} We present the first characterization of entity-level and semantic reuse in KGQA workloads, showing that both initial and traversal-level entity accesses recur and are skewed, which motivates caching as a first-class optimization for KG-augmented LLM systems.
    \item {\bf KGCache Design:} We develop KGCache, a drop-in cache layer between the KGQA engine and the KG backend that serves repeated one-hop neighborhood requests from memory, compatible with both iterative ToG~\cite{TOG_01} and one-shot RoG~\cite{rog} pipelines, and complemented by a semantic-context cache that reuses retrieved KG context across similar queries.
    \item {\bf Evaluation:} We evaluate KGCache across three cache policies, two LLMs, two KG backends, and two KGQA benchmarks, achieving up to $1.91\times$ KG retrieval speedup, and show that end-to-end gains are bounded by the share of runtime spent on KG retrieval, identifying higher-latency KG deployments as the regime where caching matters most.
\end{enumerate}

\section{Background and Motivation}
\label{sec:background}

\subsection{Reasons for KGQA existence}

Recent work by Laban et al. \cite{laban2025llmslostmultiturnconversation} shows that LLMs get lost in lengthy multi-turn conversations. This finding, however, does not by itself necessitate a RAG system. RAG is used to ground a model’s responses, but grounding can be achieved more simply: the answers to previous questions can be appended to the prompt, turning the exchange into a single-turn conversation. This accumulated context effectively plays the role of the retrieval corpus, which eliminates all external components and relies solely on the LLM’s parametric knowledge and the provided context. Table~\ref{tab:cost-per-question} shows the cost of such a workload, and the problem is evident. The number of tokens per question grows rapidly because the model must reprocess the full context for every question, and that context gets longer with each question asked.

\begin{table}[htpb]
\centering
\caption{Cost per question on WebQSP for first 300 questions.}
\label{tab:cost-per-question}
\begin{tabular}{lrrr}
\toprule
Model & LLM Cache Avg.\ tok/q & KGCache Avg.\ tok/q & Ratio \\
\midrule
ROG & 85{,}222 & 1{,}015 & 84$\times$ \\
TOG & 5{,}392  & 168     & 32$\times$ \\
\bottomrule
\end{tabular}
\end{table}

\subsection{KGQA Task and Execution Paradigms}
\label{sec:background-paradigms}

A knowledge graph is $G=(\mathcal{E},\mathcal{R},\mathcal{T})$, where $\mathcal{T} \subseteq \mathcal{E}\times\mathcal{R}\times\mathcal{E}$ is a set of (subject, relation, object) triples over entity set $\mathcal{E}$ and relation set $\mathcal{R}$. Given a question $q$ with topic entities $\mathcal{E}_q \subset \mathcal{E}$, a KGQA system retrieves a context $C_q$, a verbalized subset of $\mathcal{T}$ relevant to $q$, and generates an answer $a_q = \mathrm{LLM}(q, C_q)$. Retrieving $C_q$ requires one or more accesses to the KG backend. 

While there are several approaches for KGQA systems, we study two popular paradigms.
RoG is a planning-retrieval-reasoning framework~\cite{rog}. The LLM first generates a relation-path plan $P_q$ grounded in $\mathcal{R}$, without accessing the KG. The plan is then executed against the KG in a single pass to retrieve $C_q$, and the LLM reasons over $C_q$ to produce the answer. 
ToG instead retrieves $C_q$ iteratively~\cite{TOG_01}. 
Starting from a frontier $F_0 \subset \mathcal{E}$, at each hop $t$ the system queries the backend for the neighborhoods of $F_{t-1}$, and the LLM prunes the result to a smaller frontier $F_t$. This repeats up to a fixed depth, and $C_q$ is formed from the triples visited along the way. Because each pruning step is explicit, ToG offers knowledge traceability and correctability that one-shot planning does not.

Both paradigms access KG by requesting the one-hop neighborhood of an entity, issued repeatedly across hops (ToG) or path executions (RoG). 
If the same entity is requested more than once, or if different questions retrieve overlapping context, this access is redundant and can potentially be served from a cache instead of the backend. 
This offers two optimization opportunities:
\begin{itemize}
\item {\bf Entity Reuse:} How often do entity-level KG requests repeat, within and across questions? 
\item {\bf Semantic Reuse:} Do semantically similar queries retrieve overlapping KG contexts? 
\end{itemize}

\subsection{Evidence for Entity-Level Reuse}

\label{sec:motivation_caching}

\begin{figure*}[htpb]
    \centering
    \includegraphics[width=\textwidth]{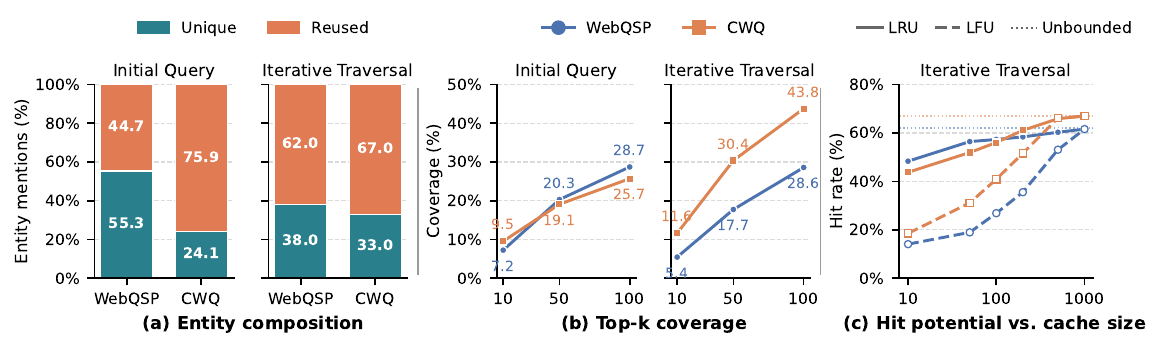}
    \caption{Entity reuse and frequent-entity coverage in KGQA workloads: 
    (a) Unique vs. reused entities in initial/iterative ToG.
    (b) Coverage of top-$K$ most accessed entities.
    (c) Cache hit potential.}
    \label{fig:entity-characterization}
\end{figure*}

Figure~\ref{fig:entity-characterization}a characterizes entity reuse at two levels: initial query entities and traversal-level entities. We label an occurrence as \emph{reused} when its entity identifier appears more than once in the corresponding analyzed trace. 
This aggregate metric does not distinguish within-question from cross-question recurrence. All other occurrences are labeled \emph{unique}. The first plot considers only the initial entities extracted from each question. 
In WebQSP, $44.7\%$ of initial-entity occurrences correspond to reused entities, while in CWQ this fraction is $75.9\%$.
The second plot extends the analysis to all entities reached during iterative ToG traversal. 
For WebQSP, reused entities account for $62.0\%$ of traversal-level entity occurrences, while for CWQ they account for $67.0\%$. 
These results show that reuse remains strong both for the initial entities explicitly mentioned in the question and for the entities explored along the reasoning path.
Such reuse creates a natural opportunity for caching, since the one-hop neighborhoods retrieved for these entities are likely to be requested again. 

Figure~\ref{fig:entity-characterization}b further examines whether entity reuse is concentrated among a small set of frequently accessed entities. 
The first plot considers only the initial entities extracted from each question. For WebQSP, the top-10, top-50, and top-100 initial entities cover $7.2\%$, $20.3\%$, and $28.7\%$ of all starting entity occurrences, respectively. 
The second plot extends this analysis to all entities reached during iterative ToG traversal. 
 For WebQSP, the top-10, top-50, and top-100 traversal entities cover $5.4\%$, $17.7\%$, and $28.6\%$ of all iterative entity occurrences, respectively. 
CWQ also shows similar trends.
These results show that entity access is skewed toward a relatively small set of entities rather than being uniformly distributed across the workload.

The reuse percentage and top-k coverage characterize how frequently entities recur across the workload, but they do not indicate whether a repeated entity remains resident in a finite cache until its next access. To capture this temporal effect, Figure~\ref{fig:entity-characterization}c additionally reports cache hit potential using bounded LRU and LFU caches across multiple capacities. Across the evaluated cache sizes, cache hit potential ranges from $\approx18-\approx62\%$ on WebQSP and $\approx20-\approx67\%$ on CWQ, showing that a substantial portion of the observed reuse occurs within reusable temporal windows.


\subsection{Evidence for Semantic-Level Reuse} 
\label{sec:motivation_semantic}

In this section, we analyze queries judged similar by their embedding score to determine whether they can reuse KG context and how closely the reused and normally retrieved contexts overlap at the entity level. Embedding similarity is only a retrieval signal and does not by itself establish that two questions have identical intent. Unlike the approach of Dominic et al.~\cite{other_caching}, we cache the KG context provided to the LLM rather than the LLM's full response. 
For an incoming query, the cache selects an existing element with the highest cosine similarity. 
A semantic hit occurs if this score meets the threshold $\tau \in [0,1]$.
Semantic caching can increase hit rates by allowing similar, rather than only identical, queries to reuse cached contexts, but it adds embedding metadata and similarity-search overhead. Lowering $\tau$ admits more hits but increases the risk that the reused context differs from the context produced by normal retrieval. For each semantic hit, we compare the entity identifiers in the cached context with those in the normal context for the incoming query and average the resulting overlap score over hits.

\begin{figure*}[htbp]
    \centering
    \includegraphics[width=0.9\textwidth]{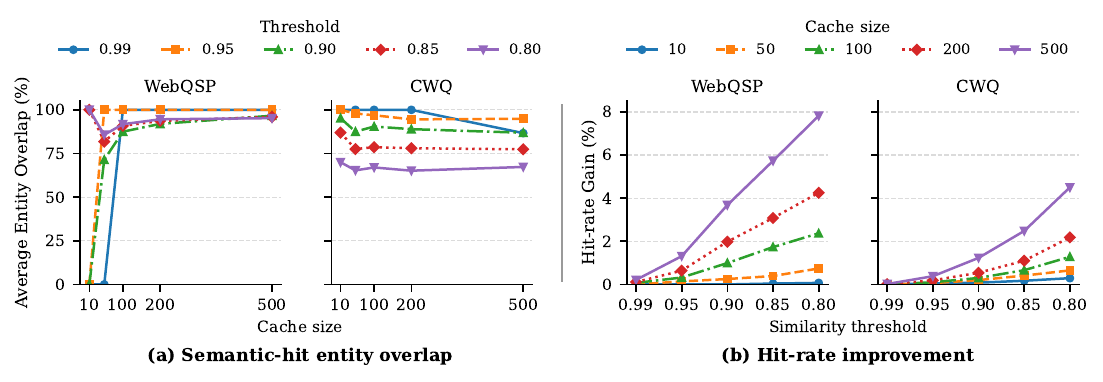}
    \caption{Effectiveness of semantic caching in KGQA workloads.
    (a) Average entity-set overlap for hits; $0\%$ indicates configurations with no semantic hits.
    (b) Hit-rate gain over exact query matching.}
    \label{fig:overlap_hitrate}
\end{figure*}

Figure~\ref{fig:overlap_hitrate}a shows the average overlap for each cache size and threshold. For WebQSP, overlap below $100\%$ first appears at $\tau=0.90$ among configurations with hits, whereas CWQ exhibits incomplete overlap at every evaluated threshold. Figure~\ref{fig:overlap_hitrate}b shows the percentage-point gain compared to exact matches for the same cache size. At $\tau=0.90$, WebQSP's gain is $4\%$ on a 500-entry cache. However, this does not establish that $\tau=0.90$ is F1-neutral.
Because CWQ exhibits incomplete overlap at every evaluated threshold with semantic hits, we use the stricter $\tau = 0.95$ for CWQ and $\tau = 0.90$ for WebQSP; Section~\ref{sec:result-sim-cache} shows that F1 remains within 1.2 percentage points of the no-cache baseline at these thresholds. Prefix caching is complementary rather than a substitute: it lowers the cost of reprocessing LLM input, while KGCache avoids backend accesses across standalone questions and sessions. Within a ToG question, the LLM context holds only pruned evidence, whereas KGCache stores the full neighborhood $N(e)$ that later hops may request.

\section{Methods}

\subsection{Cache Design}
\label{sec:cache-design}
To minimize redundant graph traversals, we position an in-memory cache between the KGQA engine and KG backend.
The cache stores one-hop KG neighborhoods at entity granularity under the assumption of a static KG snapshot.
We implement the cache as a hash map keyed by entity identifier $e$. Its value $N(e)$ holds the complete, unpruned one-hop neighborhood, including all incoming and outgoing $(\text{subject}, \text{relation}, \text{object})$ triples required for traversal. 
Because $N(e)$ is cached before question-specific LLM pruning, repeated requests for $e$ reuse the same cached graph structure.

{\bf Access Policy \& Operations.} When the KGQA engine requests $N(e)$, the cache executes the following policy:
\begin{enumerate}
    \item Lookup \& Hit: The system queries the hash map for key $e$. On a hit, it returns the one-hop neighbors $N(e)$ immediately, bypassing the KG backend.
    \item Miss \& Insertion: On a miss, the system fetches $N(e)$ from the backend (e.g., a graph database), materializes the neighborhood object, and inserts $(e, N(e))$ into the hash according to the replacement policy discussed next.
    \item Eviction: If the cache exceeds its capacity of $C$ entity entries, an eviction policy selects and removes an entry based on the replacement policy.
\end{enumerate}

{\bf Complexity \& Overhead.} Let $C$ be the number of cached entities and $|N(e)|$ the number of triples in an entry. Key lookup and insertion take $O(1)$ average time. Returning a cached reference takes $O(1)$ time, while copying or serializing a neighborhood requires $O(\vert{}N(e)\vert{})$ time. 
A miss additionally incurs backend query latency. Metadata and keys require $O(C)$ memory, while payload storage consumes $O(\sum_{e \in \mathrm{cache}} \vert{}N(e)\vert{})$ space. In our experiments, the capacity is bounded by the entity count $C$ (counting each neighborhood as one entry) to isolate the dynamics of reuse at the entity-level, although the byte sizes vary with $\vert{}N(e)\vert{}$.

\subsection{Cache Management Policies}
\label{sec:cache-management}
\begin{figure}[!t]
  \centering
  \includegraphics[width=0.6\columnwidth]{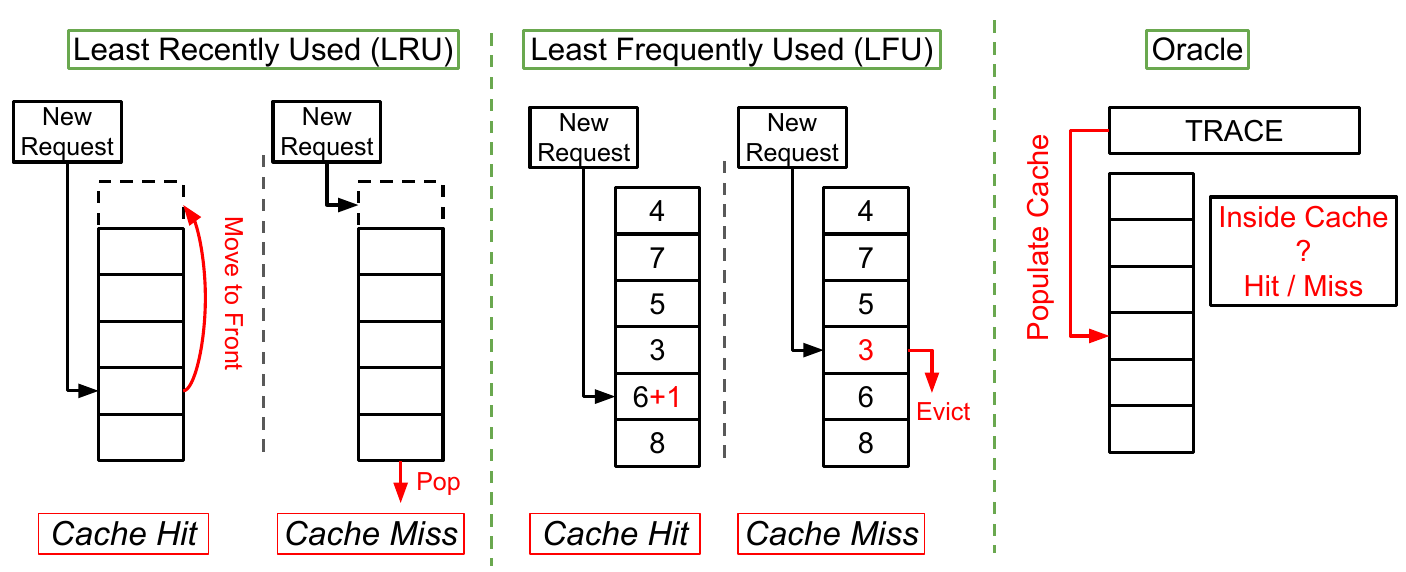}
  \caption{Overview of KG neighborhood cache policies:
LRU uses temporal locality, moving hits to the front and evicting least-recently-used entries.
LFU uses frequency locality, incrementing access counts on hits and evicting least-frequently-used entries.
Trace-aware static-frequency leverages the full request trace for global popularity but isn’t a theoretical upper bound.}
\label{fig:cache-design}
\end{figure}

The following policies determine which entry is evicted when a miss occurs and the cache is full. Figure~\ref{fig:cache-design} illustrates the lookup and replacement process. We evaluate three cache-management policies. The first policy is \emph{Least Recently Used} (LRU), which exploits temporal locality. A standard hash table with a recency list allows a hit to move its entry to the most-recent position and a full-cache miss to evict the least-recent entry in $O(1)$ average time. The second policy is \emph{Least Frequently Used} (LFU), which exploits frequency locality. LFU maintains an access counter for each cached entity, increments it on a hit, and evicts an entry with the minimum count when a miss requires space. 
Here, counter updates take $O(1)$ average time, while selecting a victim by a direct scan takes $O(C)$ time.

Finally, we evaluate a \emph{trace-aware static-frequency} reference policy, labeled \emph{Oracle} in the results for brevity. Given the complete request trace offline, it pre-populates the cache with the $C$ most frequently requested entities and does not update entries during evaluation. This policy assumes global-frequency knowledge and is therefore used for hit-rate comparison rather than the runtime experiments. 
It is not a theoretical upper bound since LRU can outperform it when reuse is concentrated in short temporal windows (Section~\ref{sec:result-cache-hit}).

\subsection{Semantic Caching Management}
\label{sec:semantic_cache_management}
Semantic caching extends the entity-level framework to query-context caching in four steps.

\begin{enumerate}
\item Encoding: Each cache entry stores a query embedding together with the verbalized KG context constructed for that query. Embeddings are computed with \texttt{all-MiniLM-L6-v2} and L2-normalized, so cosine similarity reduces to a dot product:
\begin{equation}
    \text{sim}(d_i, d_j) = \mathbf{u} \cdot \mathbf{v} = \sum_{k=1}^{384} u_k v_k
    \label{eq:similarity_3}
\end{equation}
\item {Retrieval:} For an incoming query, the cache computes its embedding and searches all $C$ stored embeddings of dimension $d=384$ for the maximum similarity score. This exhaustive comparison costs $O(Cd)$ time, excluding computation of the query embedding itself.
\item {Hit/miss decision:} A semantic hit is triggered when the maximum score meets a tunable threshold $\tau \in [0,1]$. 
On a hit, the cache reuses the associated context, skipping KG traversal and its intermediate context-construction steps. The answer-generation LLM is still invoked for the incoming query. 
On a miss, the configured KGQA pipeline runs in full.
\item {Insertion and eviction:} A miss inserts the query embedding and resulting context as a new cache entry. Under the online policies, LRU and LFU operate on these query-context entries as described in Section~\ref{sec:cache-management}.

\end{enumerate}





Semantic caching is complementary to entity-neighborhood caching.
While entity caching reduces backend work at each reasoning hop, semantic caching reuses a previously constructed context across similar queries. 
On a semantic miss, entity caching can still reduce KG retrieval during traversal. 

\section{Implementation}
\label{sec:implementation}
We integrate the entity-neighborhood cache with ToG and use a RoG-style plan-then-execute baseline in the semantic-cache experiments. Virtuoso \cite{openlink_virtuoso} and Oxigraph \cite{oxigraph_db} serve as interchangeable KG backends.
Before loading Freebase \cite{freebase}, we retain entity-to-entity triples, English textual triples, and typed literal triples while removing non-English textual triples.
For each dataset, we then collect the Freebase entity identifiers associated with its questions and retain every triple whose subject or object references one of those identifiers. The result is a dataset-specific collection of incident triples rather than a vertex-induced subgraph or the full Freebase graph. This construction reduces loading and query cost, so the reported KG latency should be interpreted for these evaluation-specific subgraphs.

The reported experiments use \texttt{Gemini 3.1 flash-lite} and \texttt{Claude Haiku 4.5}. A ToG query may trigger multiple calls because each traversed hop can include relation pruning, entity pruning, and an answer-sufficiency check, followed by answer generation. 
The exact count is path-dependent, but evaluated queries required five to seven calls in our experiments. We use WebQSP and CWQ because the workload characterization in Section~\ref{sec:background} shows different reuse concentration across their initial and traversal-level entity accesses.

We evaluate LRU, LFU, and the trace-aware static-frequency reference labeled Oracle. For the exact entity-cache hit-rate experiments, we vary cache capacity and report both the original question order and a randomized order for each policy. Separate runtime experiments execute the configured pipeline with and without caching, while the semantic answer-quality experiment uses the full WebQSP and CWQ test sets as described in Section~\ref{sec:result-sim-cache}.

\section{Results}

\subsection{Experimental Setup}

\textbf{Datasets}
We use the dataset-specific Freebase collections described in Section~\ref{sec:implementation}. Entity-cache hit-rate and runtime experiments use WebQSP and CWQ, while engine quality and semantic-cache answer quality are evaluated on the WebQSP and CWQ test sets.

\textbf{LLMs}
The LLMs used for these experiments are \texttt{Gemini 3.1 flash-lite} and \texttt{Claude Haiku 4.5}, accessed through a university-provided API platform. The LLMs are instructed to answer only from information obtained from the KG and not to fall back on parametric knowledge. Otherwise, a model could mask degradation in cached KG context by answering from parametric knowledge. External grounding is disabled for the same reason.

\textbf{Performance Metrics}
The primary performance metric for entity caching is KG retrieval speedup, defined as no-cache KG time divided by cached KG time. We also report cache hit rate and KGQA answer quality. For semantic-context caching, which can skip intermediate LLM calls, we report full-system speedup. KGQA metrics are calculated using the scripts provided by the dataset authors~\cite{WebQSP}.

\subsection{Accuracy Results}
\label{sec:accuracy-results}

We first evaluate our Think-on-Graph and Reasoning-on-Graph implementations. Table~\ref{tab:engine_quality} reports Exact Match,  recall, F1, and Hits@1 on WebQSP and CWQ. The best Exact Match score is $73.15\%$ using ToG and $69.37\%$ using RoG. 
We note that our focus is primarily on reducing graph traversal overhead via caching rather than improving the accuracy of KGQA.
These results simply establish the answer quality of the evaluated engines.

\begin{table*}[htbp]
\centering
\caption{ Engine Evaluation Results. EM: exact match, R: recall,
F1: F1-score, H@1: Hits@1, vir: Virtuoso, oxi: Oxigraph,
gem: Gemini 3.1 flash-lite, hu: Haiku 4.5. (Uncached baseline)}
\label{tab:engine_quality}
\small

\setlength{\tabcolsep}{4pt}

\begin{tabular}{llrrrrlrrrr}
\hline
& \multicolumn{5}{c}{RoG} & & \multicolumn{4}{c}{ToG} \\
\cline{2-6} \cline{8-11}
Dataset & Config & EM & R & F1 & H@1 & & EM & R & F1 & H@1 \\
\hline
CWQ
& gem-vir
& 65.07 & 51.53 & 51.47 & 58.93
& \hspace{6pt}
& 56.50 & 52.00 & 51.92 & 52.00 \\
\hline
WebQSP
& gem-oxi
& 68.89 & 47.12 & 48.37 & 63.68
& \hspace{6pt}
& 72.42 & 45.04 & 49.37 & 65.71 \\
WebQSP
& gem-vir
& 69.37 & 47.67 & 48.75 & 64.94
& \hspace{6pt}
& 73.15 & 45.34 & 49.60 & 66.26 \\
WebQSP
& hu-oxi
& 53.07 & 30.83 & 31.49 & 42.44
& \hspace{6pt}
& 63.82 & 42.35 & 39.72 & 61.13 \\
WebQSP
& hu-vir
& 54.12 & 30.71 & 31.56 & 42.57
& \hspace{6pt}
& 63.58 & 41.41 & 40.09 & 61.13 \\
\hline
\end{tabular}
\end{table*}

\subsection{Cache Hit Rate }

\label{sec:result-cache-hit}

\begin{figure}[htbp]
    \centering
    \includegraphics[width=0.9\textwidth]{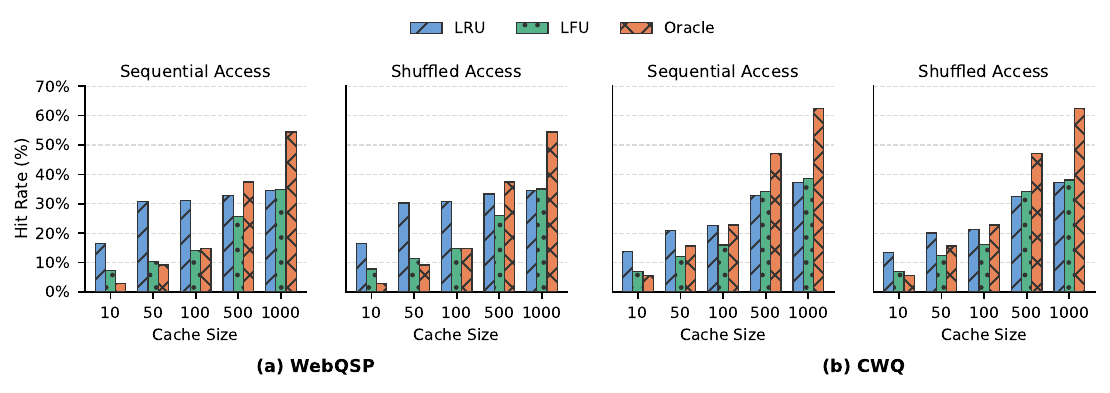}
    \caption{Cache hit rates for sequential and shuffled access orders in KGQA.
    (a) WebQSP.
    (b) CWQ.}
    \label{fig:hitrate}

\end{figure}

Figures~\ref{fig:hitrate}a and~\ref{fig:hitrate}b show the cache hit rate of ToG on WebQSP and CWQ, respectively. 
The $x$-axis shows the cache size, measured as the number of entity neighborhoods that can be stored in the cache. The $y$-axis shows the cache hit rate, defined as the fraction of KG neighborhood requests served directly from the cache. 
We evaluate three cache policies: LRU, LFU, and the trace-aware static Oracle. 
We report results under both sequential access, where questions are processed in their original dataset order, and shuffled access, where the question order is randomized.

Figure~\ref{fig:hitrate}a shows that WebQSP exhibits meaningful reuse even with small cache sizes. Under sequential access, LRU achieves a $16.6\%$ hit rate with only 10 cached entities and improves to $30.7\%$ with 50 cached entities. Increasing the cache beyond 50 entities provides smaller gains for LRU, which reaches $31.0\%$, $32.9\%$, and $34.6\%$ at cache sizes 100, 500, and 1000, respectively. LFU is less effective at small cache sizes, achieving only $7.3\%$, $10.4\%$, and $14.1\%$ hit rate at cache sizes 10, 50, and 100. 
This indicates that small-cache performance in WebQSP is driven more by temporal locality than by long-term frequency alone.

The Oracle performs poorly at small cache sizes on WebQSP, with only $2.9\%$ hitrate at size 10 and $14.8\%$ at cache size 100. However, as cache capacity increases, Oracle improves substantially, reaching $37.4\%$ at 500 and $54.4\%$ at 1000, showing that globally frequent entities become useful only when the cache is large enough. At small cache sizes, LRU outperforms Oracle because it does not adapt to short-term bursts of reuse. Shuffled-access results follow the same trend as sequential access, showing that reuse in WebQSP is not only an artifact of the original question ordering.

Figure~\ref{fig:hitrate}b shows that CWQ follows the same overall pattern as WebQSP, but with stronger gains at larger cache sizes. Under sequential access, LRU improves from $13.8\%$ at cache size 10 to $37.2\%$ at cache size 1000, showing that CWQ also contains substantial temporal reuse. LFU starts lower, achieving only $6.9\%$ at cache size 10, but becomes competitive as the cache grows and slightly exceeds LRU at cache size 1000 with a $38.6\%$ hit rate. The trace-aware static Oracle is also weak at small cache sizes, reaching only $5.6\%$ at cache size 10, but improves sharply to $47.2\%$ and $62.3\%$ at cache sizes 500 and 1000, respectively. This shows that CWQ contains both short-term temporal locality and strong global frequency skew. LRU captures the former effectively at small cache sizes, while LFU and Oracle benefit more as the cache becomes large enough to retain frequently reused entities. The shuffled-access results closely match the sequential-access results across all policies, suggesting that CWQ cacheability is not an artifact of the original question order.

Overall, these results show that both WebQSP and CWQ provide non-trivial opportunities for KG neighborhood caching. LRU is the strongest online policy at small cache sizes because it captures short-term temporal locality. LFU is weaker at small capacities but becomes competitive as the cache grows, especially on CWQ. The trace-aware static Oracle performs poorly at small cache sizes but improves sharply at larger capacities, showing that many repeated accesses are associated with globally frequent entities. The similarity between sequential and shuffled access further suggests that the measured cache benefits are not merely artifacts of dataset ordering.

\subsection{KG Retrieval Impact of Entity Caching}
\label{sec:result-end-to-end}
We evaluate KG one-hop neighborhood caching at cache size 1000 and compare the no-cache baseline with LRU and LFU on WebQSP and CWQ. The reported times are cumulative over 400 inputs from each dataset. Caching reduces KG retrieval time for both datasets, but the magnitude of the speedup differs even though the cache hit rates are similar. The results are shown in Table \ref{tab:kg_timing_sequential_1000}. On WebQSP, LFU achieves a $34.8\%$ hit rate and reduces KG time from $57.0\,\mathrm{s}$ to $42.3\,\mathrm{s}$, saving $14.7\,\mathrm{s}$ and providing a $1.35\times$ KG speedup. On CWQ, LFU achieves a comparable hit rate of $38.6\%$, but reduces KG time from $57.6\,\mathrm{s}$ to $30.2\,\mathrm{s}$, saving approximately $27.5\,\mathrm{s}$ and providing a larger $1.91\times$ KG speedup. This gap shows that runtime improvement depends both on cache hit rate and the cost of the avoided KG requests.


\begin{table}[htbp]
\centering
\caption{KG Time and Speedup by Cache Policy (sequential access, cache capacity 1000)}
\label{tab:kg_timing_sequential_1000}
\begin{tabular}{llrrrrr}
\hline
\textbf{Dataset} & \textbf{Policy} & \textbf{Hit Rate} & \textbf{No Cache (s)} & \textbf{Cached (s)} & \textbf{Saved (s)} & \textbf{KG Speedup} \\
\hline
WebQSP   & LRU      & 34.6\% & 57.0 & 42.3 & 14.7 & 1.35$\times$ \\
         & LFU      & 34.8\% & 57.0 & 42.3 & 14.7 & 1.35$\times$ \\
         & Oracle  & 54.4\% & 57.0 & 27.5 & 29.6 & 2.08$\times$ \\
\hline
CWQ      & LRU      & 37.2\% & 57.6 & 31.2 & 26.4 & 1.85$\times$ \\
         & LFU      & 38.6\% & 57.6 & 30.2 & 27.5 & 1.91$\times$ \\
         & Oracle  & 62.3\% & 57.6 & 16.1 & 41.5 & 3.57$\times$ \\
\hline
\end{tabular}
\end{table}
We note that entity-neighborhood caching accelerates KG lookups but does not change the LLM call sequence. 
Hence, we report KG-component speedups in this section.
Semantic-context caching, evaluated next, can reuse a completed context and skip graph traversal and its intermediate LLM calls.
We therefore report full-system speedup for semantic caching.
Because entity caching accelerates only the KG component, its end-to-end benefit is bounded by KG retrieval's share of runtime, which is small in our local, subgraph-based setup. This share, and hence the benefit, grows with full-scale KGs~\cite{cypherbench}, remote KG services, faster LLMs, and retrieval-intensive pipelines~\cite{gnn-rag} \cite{rje}.


\subsection{Semantic Context Cache Results}
\label{sec:result-sim-cache}
In this section, the results of Semantic Context Cache are analyzed. The results are analyzed on the same cache replacement policies mentioned in the previous section (LRU, LFU, Oracle).
Unlike Dominic et al.~\cite{other_caching}, who cache complete responses on a custom dataset, we cache retrieved KG context and still invoke the answer-generation LLM on both WebQSP and CWQ datasets. Table~\ref{tab:hitrate} reports cold-cache, first-pass hit rates on both WebQSP and CWQ. Exact query matching produces no hits because no question string repeats, whereas the semantic policies reach up to $8.2\%$.

\begin{table}[htbp]
\centering
\caption{Measured first-pass semantic cache hit rate in live runs (\%). G/H: Gemini/Haiku; O/V: Oxigraph/Virtuoso.}
\label{tab:hitrate}
\small
\setlength{\tabcolsep}{3pt}

\begin{tabular}{lrr|rrrr|rrrr}
\hline
& \multicolumn{2}{c|}{CWQ}
& \multicolumn{4}{c|}{WebQSP--RoG}
& \multicolumn{4}{c}{WebQSP--ToG} \\
\cline{2-11}
\textbf{Policy}
& \textbf{RoG} & \textbf{ToG}
& \textbf{G(O)} & \textbf{G(V)} & \textbf{H(O)} & \textbf{H(V)}
& \textbf{G(O)} & \textbf{G(V)} & \textbf{H(O)} & \textbf{H(V)} \\
\hline
Exact  & 0.0 & 0.0  & 0.0 & 0.0 & 0.0 & 0.0 & 0.0 & 0.0 & 0.0 & 0.0 \\
LRU    & 2.4 & 2.5 & 7.3 & 7.7 & 7.4 & 7.7 & 8.2 & 8.2 & 8.2 & 8.2 \\
LFU    & 2.4 & 2.5 & 7.7 & 7.7 & 7.7 & 7.7 & 8.2 & 8.2 & 8.2 & 8.2 \\
Oracle & 1.7 & 2.0 & 6.8 & 6.8 & 6.8 & 6.8 & 7.3 & 7.3 & 7.3 & 7.3 \\
\hline
\end{tabular}

\end{table}

\begin{figure}[htpb]
\begin{subfigure}[b]{0.47\columnwidth}
        \centering
        \includegraphics[width=\columnwidth]{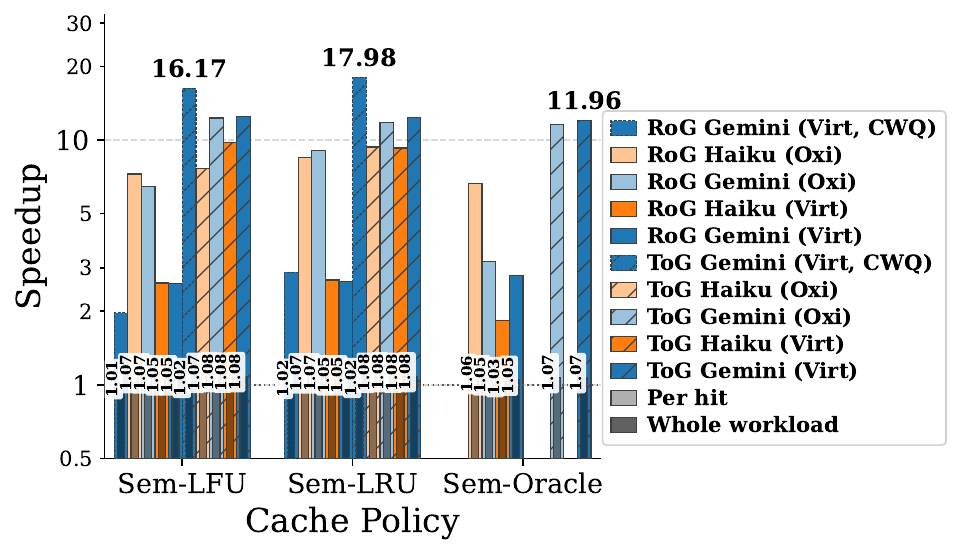}
          \caption{Cold-cache speedup of semantic-context caching on WebQSP and CWQ}.
        \label{fig:speedup}
    \end{subfigure}
    \hfill
    \begin{subfigure}[b]{0.47\columnwidth}
        \includegraphics[width=\columnwidth]{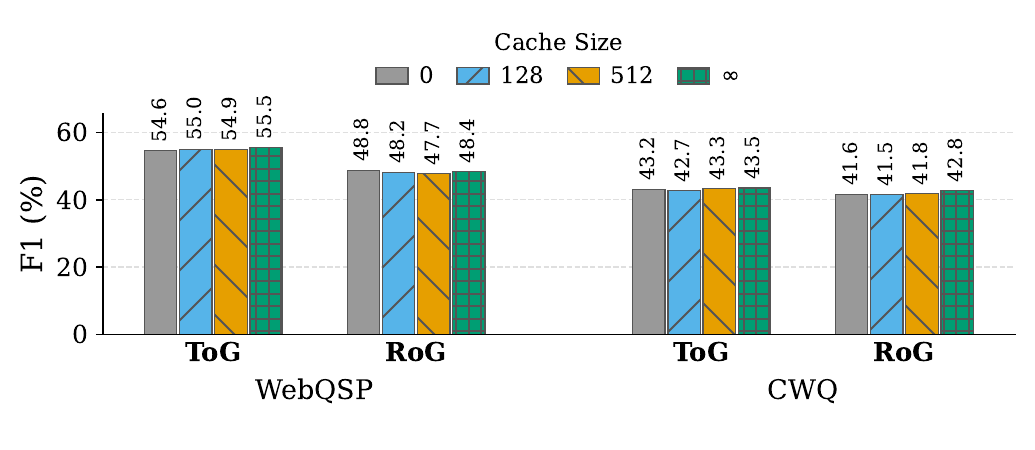}
          \caption{WebQSP F1 at $\tau=0.90$ and CWQ F1 at $\tau=0.95$ with Gemini 3.1 flash-lite across cache sizes; size 0 is the no-cache baseline using LFU policy.}
        \label{fig:cache_vs_acc}
    \end{subfigure}
  \label{fig:semantic_results}
  \caption{Semantic caching results}
\end{figure}

The KG backends run in local Virtuoso/Oxigraph containers, while the cache is an in-memory dictionary. Figure~\ref{fig:speedup} reports cold-cache full-system and per-hit speedups on WebQSP and CWQ over various configurations: each combines a KGQA engine (RoG or ToG), an LLM (Gemini or Haiku), and a KG backend (Virtuoso or Oxigraph). Exact matching remains at $1.00\times$ because it has no hits, while the semantic policies reach up to $1.08\times$. Each cache hit is up to $17.98\times$ faster then a miss, but low overall speedup is a sysmptom of low cache hit rate as shown in Table \ref{tab:hitrate}.
We evaluate F1 on the full WebQSP and CWQ test sets using Gemini 3.1 flash-lite. Figure~\ref{fig:cache_vs_acc} compares cache sizes 128, 512, and an unbounded cache with the no-cache baseline. The F1 scores differ from their respective baselines by at most 1.1\% for WebQSP and 1.2\% for CWQ, with no consistent trend across cache capacity. Table \ref{tab:replicate_variation} reports the F1-score variability across three runs of identical configurations on WebQSP. These results suggest that semantic-context reuse at $\tau=0.90$ for WebQSP and $\tau=0.95$ for CWQ does not measurably degrade answer quality.
Table \ref{tab:accuracy_threshold} shows the degradation of performance due to a lax threshold. It degrades up to 4.99\% for ToG, while having significantly less of an impact on RoG (1.46\%). 
Table \ref{tab:param_vs_live} shows the performance difference between using a KGQA workload and relying solely on the LLM's parametric knowledge. In all cases, KGQA achieves better performance than the LLM alone. Once again ToG is affected more, with a boost of up to 4.64\%, while RoG reports a boost as low as 1.20\%.

\begin{table*}[htbp]
\centering
\caption{Metric variation across replicates, sensitivity to threshold changes, and parameterized versus live configurations.}
\label{tab:accuracy_results}

\setlength{\tabcolsep}{2pt}

\vspace{-5pt}

\begin{subtable}[t]{0.27\textwidth}
    \centering
    \caption{F1(\%) variation on 3 runs of the same configuration (top 400 questions).}
    \label{tab:replicate_variation}
    \scriptsize
    \begin{tabular}{@{}p{0.30\linewidth}rrr@{}}
        \toprule
        System & Mean & Std.\ Dev. & Range \\
        \midrule
        ToG & 58.62 & 0.11 & 0.22 \\
        RoG & 51.16 & 1.29 & 2.45 \\
        \bottomrule
    \end{tabular}
\end{subtable}%
\hfill
\begin{subtable}[t]{0.31\textwidth}
    \centering
    \caption{F1(\%) under threshold changes.}
    \label{tab:accuracy_threshold}
    \scriptsize
    \begin{tabular}{@{}p{0.25\linewidth}rrr@{}}
        \toprule
        System & $\tau=0.90$ & $\tau=0.50$ & $\Delta$ \\
        \midrule
        ToG & 57.80 & 52.81 & -4.99 \\
        RoG & 50.41 & 48.95 & -1.46 \\
        \bottomrule
    \end{tabular}
\end{subtable}%
\hfill
\begin{subtable}[t]{0.38\textwidth}
    \centering
    \caption{Hit@1(\%) comparison between parameterized and live configurations.}
    \label{tab:param_vs_live}
    \scriptsize
    \begin{tabular}{@{}p{0.52\linewidth}rrr@{}}
        \toprule
        Config. & Param. & Live & $\Delta$ \\
        \midrule
        ToG--Gemini (Oxi)  & 61.62 & 65.71 & +4.09 \\
        ToG--Gemini (Virt) & 61.62 & 66.26 & +4.64 \\
        RoG--Gemini (Oxi)  & 62.48 & 63.68 & +1.20 \\
        RoG--Gemini (Virt) & 63.44 & 64.94 & +1.50 \\
        \bottomrule
    \end{tabular}
\end{subtable}
\end{table*}
\section{Related Work}
\label{sec:related_works}
\textbf{Iterative Traversal}
A prominent line of work has an LLM traverse the KG hop by hop, using beam search to control the candidate set. \textit{Think-on-Graph} (ToG)~\cite{TOG_01} uses an LLM to score and prune graph traversals until an answer is reached or a depth limit is hit; its LLM calls grow with traversal depth and per-hop relation and entity pruning. \textit{Think-on-Graph 2.0} (ToG-2.0)~\cite{TOG_02} adds a pre-traversal reasoning stage intended to reduce exploration of irrelevant branches. \textit{Search-on-Graph} (SoG)~\cite{sog} simplifies the beam formulation to a \texttt{Search()} function for adaptive one-hop expansion, reducing LLM calls while remaining competitive on Freebase and Wikidata benchmarks. \textit{Plan-on-Graph} (PoG)~\cite{pog} first generates a relation-oriented plan and then uses it to guide iterative beam traversal. Without explicit cross-query caching, these methods process questions independently, so questions sharing topic or intermediate entities may repeatedly retrieve the same neighborhoods.



\textbf{One-Shot Planning}
A second family avoids iterative planning calls by having the LLM generate a relation-path plan before KG execution. \textit{Reasoning on Graphs} (RoG)~\cite{rog} fine-tunes an LLM to generate relation paths, retrieves valid paths from the KG, and verbalizes them for a separate reader LLM to produce the final answer. \textit{KARPA}~\cite{karpa} removes this fine-tuning requirement by using a training-free LLM for global path planning over relation embeddings and matching generated paths to KG structure through embedding similarity rather than exact string matching. One-shot planning reduces planning-stage LLM calls, but KG execution can still revisit entities or relation paths across questions unless results are cached.


\textbf{Subgraph Retrieval}
A third paradigm separates KG access from downstream LLM reasoning: a retriever extracts a subgraph around the question's topic entities, and the LLM reads its verbalization without further KG access. \textit{Subgraph Retrieval Enhanced Model} (SR)~\cite{subgraph_retrieval_enhanced} introduced a trainable dual-encoder retriever decoupled from the downstream reasoner, with weakly supervised pre-training on shortest paths between topic and answer entities. \textit{ReasoningLM}~\cite{reasoning_lm} adapts a PLM to perform GNN-style subgraph reasoning through structural self-attention, tuned on synthesized subgraph--question pairs. \textit{SubgraphRAG}~\cite{subgraphrag} uses a lightweight parallel MLP triple scorer to produce flexibly sized subgraphs for an unmodified LLM reader. Questions sharing topic entities or relation chains may retrieve overlapping subgraphs.
\textit{GNN-RAG} reduces LLM-guided graph traversal using a lightweight GNN retriever~\cite{gnn-rag}, \textit{RJE} limits additional graph exploration to cases where retrieved evidence is insufficient~\cite{rje}, \textit{BYOKG-RAG} combines multiple retrieval strategies~\cite{byokg-rag}, and \textit{CypherBench} studies retrieval over full-scale KGs~\cite{cypherbench}.
These works highlight the growing importance of efficient KG retrieval, while our results show that repeated graph accesses can be amortized through caching.

\textbf{Semantic and KG-Augmented Caching}
A complementary line of work caches LLM responses rather than underlying KG retrieval. The KG-Enhanced Semantic Cache embeds queries, reuses responses for sufficiently similar queries, and organizes them using KG structure~\cite{other_caching}. KGCache instead primarily caches exact entity neighborhoods at the KG backend boundary; its semantic cache reuses retrieved KG context while retaining per-query answer generation.

\section{Conclusion}
\label{sec:conclusion}

In this work, we show that KGQA workloads exhibit substantial and skewed entity reuse during iterative Think-on-Graph traversal. Motivated by this reuse, we study two complementary approaches: caching one-hop entity neighborhoods and semantic-context caching. Using standard cache policies, we show that both can reduce redundant retrieval across iterative and one-shot KGQA pipelines.

Our study also leaves several directions for future work. The current cache is an in-memory, entity-level design that stores one-hop neighborhoods and evaluates standard LRU/LFU policies. Future work could explore more adaptive replacement policies and multi-hop caching, though multi-hop reuse introduces additional challenges because different questions starting from the same entity may require different reasoning paths. In addition, KG retrieval is a small fraction of runtime in our current setup, limiting end-to-end gains despite substantial KG-side speedups. Future work should therefore study larger or higher-latency KG deployments and hybrid KG/text RAG systems, where caching could span both structured neighborhoods and retrieved passages.

\section*{Acknowledgment}
This research was funded in part by DOE grants DE-SC0022098 and DE-SC0023349 and by NSF grants PPoSS CCF 2316233 and OAC-2339607.
Thanks to TAMU AI Chat for providing a free API with production-grade LLMs used for obtaining all results. 

\bibliographystyle{unsrtnat}
\bibliography{reference}

\appendix
\section{Appendix}
\subsection{AI disclosure}
In this work, we used generative AI tools for spell checking, grammar correction, vocabulary correction, language editing, and basic coding assistance such as style alignment of plots and comments/readme files. LLM-generated code was verified and tested for correctness by the authors. We take responsibility for the final content of this work, including text, claims, or artifacts produced with the aid of generative AI.

\subsection{Result Reproduction}
The code is available in an anonymous \href{https://github.com/EricYJA/KG_cash}{Github repository}. The repository has readme files that explain most of functionalities. In order to run the experiments, the user needs to provide their own API Key to some LLM endpoint. Current system is using OpenAI API compatible endpoints. Minor changes might be needed to use some vendor not supported by this project. Current publicly available vendors are Google and OpenAI. After the results are run, the readme files provide instructions on how to reproduce charts and their output locations. Certain discrepancies between the presented results and reproduced results are expected, especially if the dataset is run partially, or a different LLM than the ones mentioned in the paper is used.





\end{document}